\documentclass{article}

\usepackage{arxiv}

\usepackage[utf8]{inputenc} 
\usepackage[T1]{fontenc}    
\usepackage{hyperref}      
\usepackage{url}           
\usepackage{booktabs}       
\usepackage{amsmath}
\usepackage{amsfonts}       
\usepackage{nicefrac}      
\usepackage{microtype}     
\usepackage{graphicx}

\usepackage{siunitx}

\renewcommand{\headeright}{Technical Note}
\renewcommand{\undertitle}{Technical Note}
\renewcommand{\shorttitle}{TeCoMiner}

\DeclareMathOperator{\pos}{pos}
\DeclareMathOperator{\Idf}{Idf}

\title{TeCoMiner: Topic Discovery Through Term Community Detection}
\date{March 18, 2021}

\author{Andreas Hamm \\ {\bf Jana Thelen} \\
            {\bf Rasmus Beckmann}\\ {\bf Simon Odrowski}\\ 
            \texttt{\{Andreas.Hamm, Rasmus.Beckmann, Simon.Odrowski}@dlr.de\}\\
	Think Tank\\
	German Aerospace Center DLR\\
	Cologne, Germany \\            
}

\begin{document}

\maketitle

\begin{abstract}
This note is a short description of TeCoMiner, an interactive tool for exploring the topic content of text collections. Unlike other topic modeling tools, TeCoMiner is not based on some generative probabilistic model but on topological considerations about co-occurrence networks of terms. We outline the methods used for identifying topics, describe the features of the tool, and sketch an application, using a corpus of policy related scientific news on environmental issues published by the European Commission over the last decade.
\end{abstract}


\section{Introduction}
The rapidly increasing amount of electronically available texts has augmented the importance of automatized unsupervised methods for text exploration and analysis. A very typical task is to identify the themes which are latent in massive text collections. Not only does this help to obtain a quick overview over the content of the text collection, but it also enables a structured analysis of relationships and developments reflected in the texts. 

In contrast to supervised text classification, the discovery of topics is quite an open-ended endeavor, leaving an important role to subject-related interpretation by domain experts. Our own experience as an interdisciplinary team involved in monitoring and assessing huge text collections has led us to leave the well-trodden path of probabilistic topic models and explore the possibilities of detecting topics as communities in term networks. We have applied this approach to several text corpora \cite{Thelen2020} (scientific publications, political texts, RSS news feeds) and found it to be advantageous with regard to the ease of topic interpretation and to the command over topic granularity.

TeCoMiner is a software tool designed for users who want to apply the {\bf Te}rm {\bf Co}mmunity approach to topic detection to their corpora of interest. Before we describe the tool and demonstrate its application we will give an overview over the underlying methods.

\section{Related Work}
During the last two decades probabilistic topic models have dominated the topical analysis of text corpora, seminally influenced by \cite{Blei2003}. Based on popular software packages and more specialized advanced methods, this has led to many applications in areas as diverse as scientometric publication analysis, social media monitoring, literature studies, historical text exploration, and the social sciences. For an extensive overview we refer to \cite{BoydGraber2017}. 

The question of how to evaluate topic quality is problematic. For the applications mentioned above, topic interpretability by humans is the key measure of success, but this is a concept which is not easy to grasp in a computational way. Various metrics of topic coherence \cite{Roeder2015} have been studied as indicators of interpretability. Recently it was suggested to involve word embeddings in the assessment of topic coherence \cite{Fang2016}. We found this approach useful in our own comparative experiments with various topic models \cite{Thelen2020}, and it has influenced the way we present topics in TeCoMiner. 

Visualization methods play an important role in aiding topic interpretation; \cite{Sievert2014, Kucher2018} are only two out of many examples designed for depicting probabilistic topic models for human consumption. TeCoMiner uses similar visualizations but is geared towards network based topic detection.

The idea that topics can be interpreted as communities in networks which embody the relations between words and documents of a corpus came up in various forms \cite{Sayyadi2013, Lancichinetti2015, Dang2018, Gerlach2018}. In these settings, topics can be found by applying one of the many methods of community detection \cite{Fortunato2016}. We will explain below how we framed this approach in a way that leads to enhanced interpretability and controlled granularity of topics.

\section{Methods}
\begin{figure}[h]
  \centering
  \includegraphics[width=0.8\linewidth]{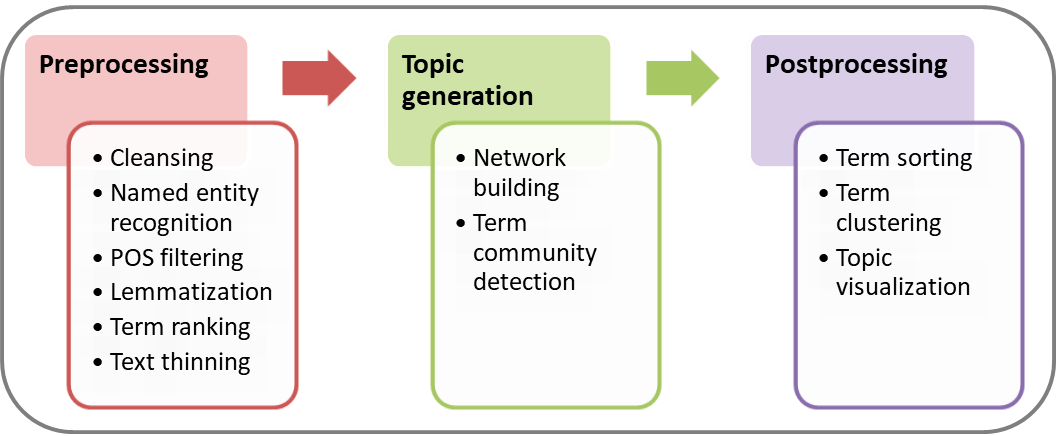}
  \caption{Topic identification pipeline in TeCoMiner}
  \label{Fig:1}
\end{figure}

While the generative approach of probabilistic topic modeling follows the sound methodology of statistical machine learning with a high potential for insights into the genesis of a corpus, it is built on certain assumptions about document generation and prior distributions that are disputable. In contrast, here we take a more phenomenological position when exploiting observed word co-occurrences in the corpus documents for topic detection. This can be done by studying the characteristics of co-occurrence networks of terms. Topics then show up as communities of strongly connected terms. However, we suggest that it needs careful pre- and postprocessing steps for achieving results that are competitive or even better compared to probabilistic topic models.
Here we briefly describe the core elements of our processing pipeline (Figure \ref{Fig:1}). More details can be found in \cite{Hamm2021b}.

\subsection{Term Ranking} \label{Sec:Ranking}
Starting from a corpus of raw texts, we first follow standard NLP preparation procedures: Removal of unwanted tokens and stopwords; POS filtering that retains only adjectives, nouns, and proper nouns; lemmatization. 

With regard to multiword expressions we experimented with various approaches. While merging words which form a unique expression is beneficial for the interpretation of term communities, the confusing effect of word combinations that just appear to be common by statistical observations without bearing a special meaning can be more harmful. We therefore consider only multiword combinations that show up in named entity recognition.

In preparation of a viable term network it is essential to reduce the number of terms retained in each document even further. For this purpose, we introduce a term ranking in each document and keep only the top-ranked terms. This is related to unsupervised keyword extraction which aims at finding those terms of a document that are most significant for its content. We have developed an approach - posIdfRank - which combines several ideas \cite{Hamm2019}: the \mbox{PageRank} inspired voting based on local word neighborhood associations introduced in \cite{Mihalcea2004}, a weighting according the absolute position in the text \cite{Florescu2017}, and counterbalancing the influence of unspecific words by the inverse document frequency \cite{Jones1972}.

Technically, we obtain the ranking values for the $n$ terms of a document as the stationary distribution of a Markov chain \mbox{$x_{t+1}=Gx_t$} on the n-dimensional space of terms with a transition matrix 
\begin{displaymath}
  G_{ij} = \alpha \frac{\Idf_j f_{ij}}{\sum_{k=1}^n \Idf_k f_{ik}} + (1-\alpha)\frac{(1+\pos_i)^\beta \Idf_i}{\sum_{k=1}^n (1+\pos_k)^\beta \Idf_k} ,
\end{displaymath}
where $\Idf_i$ is the inverse document frequency of term number $i$, $\pos_i$ is the earliest position of that term and $f_{ij}$ is the frequency how often the terms number $i$ and $j$ lie in the same neighborhood window of size $w$. $\alpha$, $\beta$, and $w$ are tuning parameters which we chose as $\alpha=0.9$, $\beta=-0.9$ and $w=11$ after experiments with documents with pre-assigned keywords \cite{Hamm2019}.

Each document is then thinned by keeping only the top $P$ percent of terms according to this ranking, where we choose $P$ between 10 and 80 depending on the average length of the corpus documents. 

\subsection{Term Community Detection} \label{Sec:Modularity}

We define a network with all the terms contained in the thinned documents as vertices $\mathcal{V}$. The edge weight $A_{ij}$ between two vertices $v_i$ and $v_j$ is defined to be the number of documents in which $v_i$ and $v_j$ appear together.
Pruning edges between rare terms is an option for large corpora with long texts.

Communities in networks are, intuitively speaking, vertex groups with strong linkage inside the groups but only loose connections to other groups. Comparing various community detection approaches we found that modularity optimization produces particularly good topics with respect to interpretability \cite{Thelen2020}. 

In TeCoMiner we use the generalized modularity definition introduced in \cite{Reichardt2006}: We call a partition $\mathcal{C} = \{C_s, s=1, \dots, m\}$ of $\mathcal{V}$ a candidate of communities. 
The generalized modularity, $\mathcal{H}_{\gamma}(\mathcal{C}) = \mathcal{I}(\mathcal{C}) - \gamma \mathcal{J}(\mathcal{C})$, compares for a candidate partition $\mathcal{C}$ the fraction of edge weights inside of candidate communities, $\mathcal{I}(\mathcal{C}) = \frac{1}{2m}\sum_{i,j} A_{ij} \delta_{c(i) c(j)}$, for the given graph on the one hand with the expected fraction of edge weights inside candidate communities for a random network with the same degree distribution, 
$\mathcal{J}(\mathcal{C}) = \frac{1}{(2m)^2}\sum_{i,j} k_i k_j \delta_{c(i) c(j)}$, on the other hand; here 
$k_i = \sum_j A_{ij}$ denotes the weighted degree (weighted number of edges) of vertex $v_i$, $m=\frac{1}{2}\sum_i k_i$ is the weighted total of edges in the network, $c(i)$ enumerates the candidate community of vertex $v_i$, and $\delta_{ij}$ is Konecker's delta.
The partition which maximizes $\mathcal{H}_{\gamma}(\mathcal{C})$ describes the optimal communities in the sense of high intra-group linkage compared to the expectations in a random situation. The parameter $\gamma$ influences how strongly one values the gain of additional intra-group edge weights. With $\gamma = 0$ one does not compare to the random situation at all and therefore the optimal solution is one all-embracing community. With $\gamma \rightarrow \infty$ intra-group links practically do not get rewarded, so that the extreme partition into one-vertex communities appears as the optimal solution. Hence, $\gamma$ can be used as a resolution parameter: smaller values lead to a few big communities, larger values to many small communities. 

Maximizing the modularity is known to be NP-hard. However, there exist efficient greedy algorithms for finding local modularity maxima. In TeCoMiner we use the Leiden algorithm \cite{Traag2019}.

\subsection{Term Community Presentation} \label{Sec:Wordcloud}

Community detection in the term co-occurrence network partitions all terms into topics. Consequently, a topic is typically characterized by some hundreds or even thousands of terms. Intrinsically, this list of terms does not come with any order. This is different from the situation in probabilistic topic models where the model produces a probability distribution over topic words.

Therefore, we introduce two criteria of how to structure the topic terms when presenting them for human interpretation: a rating of terms for sorting them according to significance, and a stratification of terms into layers of semantically similar terms.

In subsection \ref{Sec:Ranking} we have already introduced posIdfRank as a method for finding the most significant terms per document. Now we need a measure for significance within the whole corpus. A naive average over the posIdfRanks of a term from all documents in which it occurs would  be unfair because of the very different document frequencies of terms. A more appropriate way of rating is the Bayesian average as known from scoring systems. Concretely, we rank a term $a$ by calculating $x(a) = \frac{0.3 C + s(a)}{C + d(a)}$ where $d(a)$ is the number of documents containing $a$, $s(a)=\sum_u{x_u(a)}$ with $x_u(a)$ equal to 3, 2 or 1 depending on whether $a$ is among the top 5, 10 or 15 percent of terms in document $u$ according to posIdfRank, and $C$  is the sum of the number of all unique terms per document devided by the number of unique terms in the whole corpus, see \cite{Thelen2020} and \cite{Hamm2021b}.

We discover semantically similar terms by mapping all topic terms into a 300-dimensional vector space with a pre-trained fastText word embedding \cite{Bojanowski2017}. In this space we form groups of semantically related terms by agglomerative clustering based on Euclidean distance.

We use these two structuring principles for an easy-to-grasp visualization of topic terms in the form of a stratified word cloud which shows the terms in sizes according to their significance ranking positions and in colored horizontal strata which bring together semantically related terms.

\section{Demonstration}

We discuss how users can readily apply term community topic detection based on the methods described in the previous section with TeCoMiner, a software tool we wrote in Python utilizing in particular the packages pandas, spaCy, python-igraph, scikit-learn, gensim, wordcloud, Bokeh, HoloViews, and Panel. Here we briefly describe its features, which were developed in close cooperation between data scientists and end users, and introduce our demonstration case.
 
\subsection{TeCoMiner Web Application}

TeCoMiner runs as a single-page application in a web browser. There are two work phases: first, uploading and preprocessing of corpora through the {\bf Add corpus} feature (with an option to change the parameters $\alpha$, $\beta$, $w$ and $P$ mentioned in subsection \ref{Sec:Ranking}), second, interactive topic detection and analysis within those corpora.
Preprocessed corpora can be looked at in several views presented as tabs: {\em Model}, {\em Topic}, {\em Document}, and, depending on the corpus, further tabs for metadata connected to the  documents\textemdash in the present example the tabs {\em Time} and {\em Theme}. 

\begin{figure}[h]
  \centering
  \includegraphics[width=\linewidth]{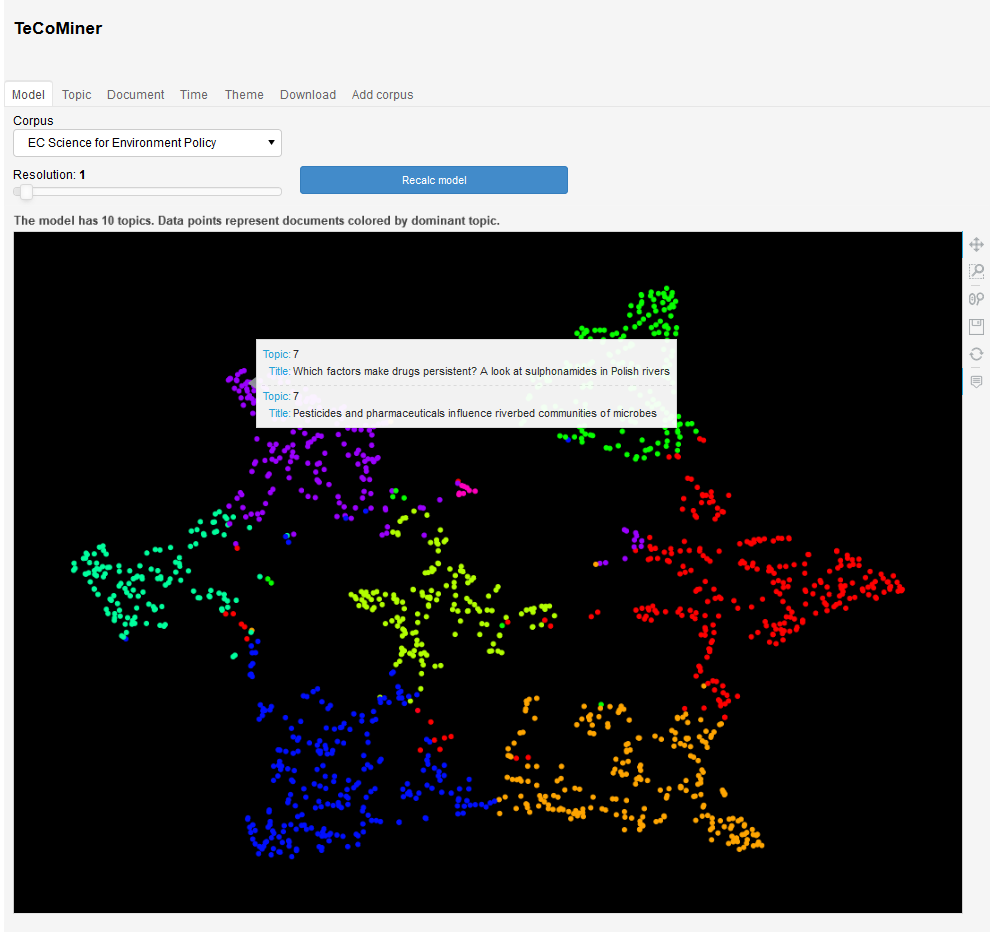}
  \caption{TeCoMiner Model tab, with a graphical overview of the model and the option to recalculate models with varying resolutions. Here: view of a resolution-1 model of the EU Science for Environment Policy News Alert 2011–2020 corpus}
  \label{Fig:2}
\end{figure}

The {\bf Model tab} (Figure \ref{Fig:2}) provides an overview of the current community topic model in the form of a two-dimensional t-SNE plot \cite{vanDerMaaten2008}. The dots represent documents; in a model with $N$
topics they are first placed in an $N$-dimensional space according to the proportions each topic contributes to the document and are then mapped to two dimensions via t-SNE. The color of each dot is chosen depending on the topic with highest proportion in that document. The title of each document can be displayed through mouse-over. Large unicolored clusters represent the most dominant topics; scanning the titles involved gives a first impression of what those topics deal with. Dense clusters hint at sharp topics. Coalescing clusters indicate topical relationships.

On the Model tab it is also possible to generate a new topic model after choosing a value for the resolution parameter $\gamma$ (see subsection \ref{Sec:Modularity}).

Topics can be analyzed in more detail by selecting TeCoMiner’s {\bf Topic tab} (Figure \ref{Fig:3}). For a topic chosen from a drop-down list, the left side of the screen shows a stratified word cloud of the topic terms as described in subsection \ref{Sec:Wordcloud}. With this form of presentation the user can recognize at a glance the subject-related common ground in about 100 terms. Highly ranked words stand out by size, and the colored horizontal strata group related terms.

On the right side of the screen, there is a list of up to 30 documents in which the selected topic takes a proportion of more than 15 percent, sorting the documents with highest proportion to the top. 

\begin{figure}[h]
  \centering
  \includegraphics[width=\linewidth]{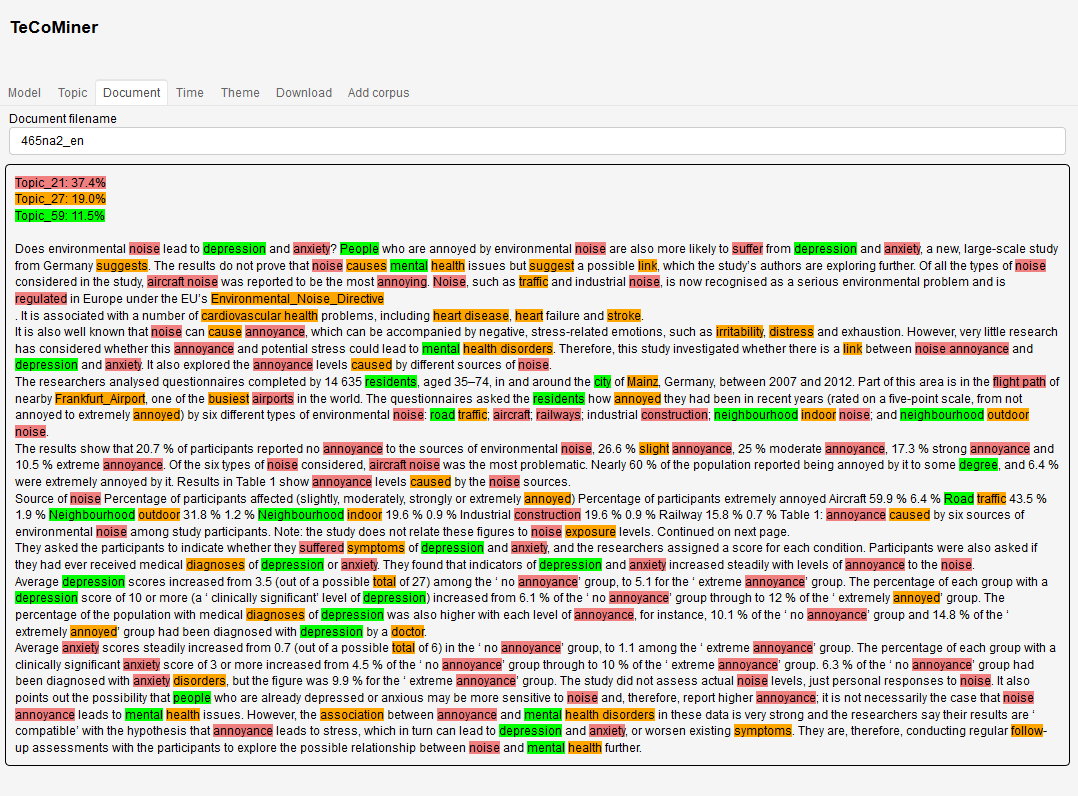}
  \caption{Document tab, showing highlighted topic terms within a document. Here: A document with title ``Does environmental noise lead to depression and anxiety?'' Three topics have a share of more than 10 percent in this document: \textit{noise polution} (Topic 21; red), \textit{health issues} (Topic 27; orange), and \textit{urban living conditions} (Topic 59; green).}
  \label{Fig:4}
\end{figure}

The {\bf Document tab} (Figure \ref{Fig:4}) shows the full text of a document and highlights in it all terms that belong to any topic with a proportion of at least 10 percent. Different topics are marked in different colors. Documents can be selected either by entering their file name or by opening them from the document list of the Topic tab. On the one hand, this tab is particularly useful for understanding and interpreting topics if word clusters and document titles have not been sufficiently insightful. On the other hand, it also provides deeper insights into how different topics relate to each other. 

With the {\bf Time tab} it is possible to follow the temporal evolution of topics. This tab presents as line charts the time series of the accumulated proportion which the (multi-)selected topics have in all the documents published in each single month. 

The {\bf Theme tab} refers to pre-assigned thematic tags (like ``Biodiversity'', ``Climate change'', and ``Sustainable mobility'') which are provided with the documents of the present corpus. It visualizes the connection between these tags and the detected topics, which is useful as a consistency check and as support for topic interpretation.

Next to the interactive features, on the {\bf Download tab} TeCoMiner offers an Excel export of the stratified topic term clouds and the topic distribution per document.

\subsection{Application Case}

We show some results derived from a collection of \num{1463} articles scraped from the European Commission's {\em EU Science for Environment Policy News Alert}\footnote{https://ec.europa.eu/environment/integration/research/newsalert/} from April 2011 until May 2020. They summarize environmental research studies for a political audience or decision-makers in general, therefore using a non-technical and accessible language. The article length varies from 100 to \mbox{1400 words}.

Preprocessing including thinning with $P=33.3\%$ of that corpus takes hardly more than 10 minutes on an Intel Core i7 PC. This produces a term network of \num{16152} vertices and \num{1149078} edges. Calculation of a term community topic model including visualizations runs in 20 seconds so that users can easily explore topics of variable granularity in an interactive manner.

The majority of topics produced by TeCoMiner are readily interpretable with the help of the visualizations provided. The only exceptions are some topics that just consist of very few terms; these are outliers in the term co-occurrence network, appearing only in a small number of documents. Such topics are easily recognizable so that they can be ignored in the analysis.

While the lowest resolution parameter $\gamma=0.8$ leads to four very broad topics, which can be described as {\em Sustainable technologies and policies}, {\em Pollution and emissions}, {\em Aquatic ecosystems}, and {\em Wildlife and farming}, at highest resolution, $\gamma= 2.5$, one can recognize 58 identifiable topics. 

\begin{figure}[h]
  \centering
  \includegraphics[width=\linewidth]{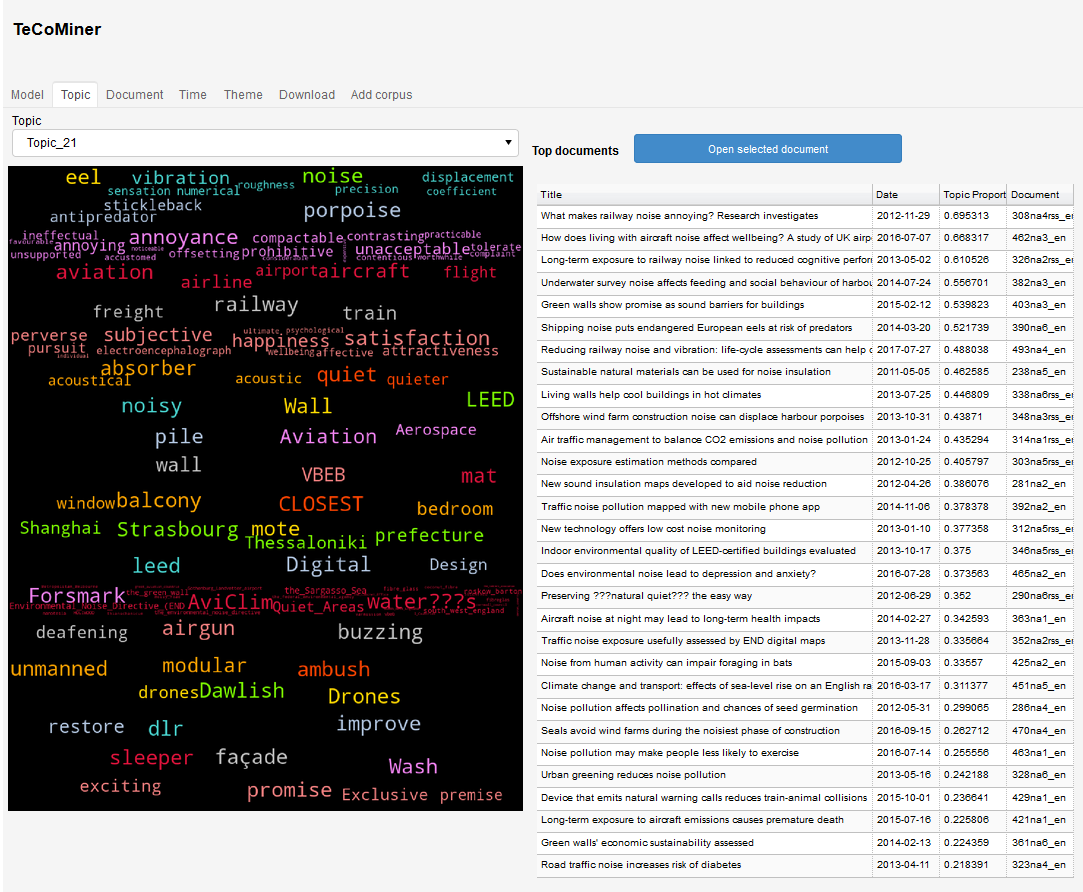}
  \caption{Topic tab, showing stratified word cloud of topic terms and list of representative documents. Here: view of a topic in a resolution-2.2 model which can be interpreted as ``noise pollution''}
  \label{Fig:3}
\end{figure}

Users can interpret the topics based on three pieces of information: the prominent topic terms visible in the stratified word cloud, the document titles of the connected documents (see Figure \ref{Fig:3}), and the context of the topic terms as depicted in the highlighted phrases on the document tab.

By changing the resolution parameter, users can determine the level of granularity with which they want to scan through the corpus. For instance, within the broad topical area of {\em Pollution and emissions} many more specific topics like {\em Air pollution}, {\em Light pollution}, {\em Noise pollution} (see Figure \ref{Fig:3}), {\em Pharmaceutical pollution}, {\em Ship emissions}, etc., can be recognized with increasing resolution. 

TeCoMiner gives decision-makers a coherent overview about the topic structure in a large document data set in a level of detail of their choice. Also, the highlighted documents facilitate speed-reading of representative example texts. Analysts can recognize trends and events in the time series and use the tool for investigating topical interrelations.

\subsection*{Acknowledgement}
We are grateful to Mark Azzam for stimulating discussions.

\bibliographystyle{unsrt}
\bibliography{TeCoWOD}

\end{document}